\documentclass[]{spie}  

\usepackage{comment}
\usepackage{booktabs}
\usepackage{seqsplit}
\usepackage{url}
\usepackage{amsmath,amsfonts,amssymb}
\usepackage{graphicx}
\usepackage{gensymb}
\usepackage{rotating}  
\usepackage{caption}
\usepackage{subcaption}
\usepackage{enumitem}
\usepackage{xcolor}
\usepackage{float}
\usepackage[colorlinks=true, allcolors=blue]{hyperref}
\usepackage{listings}

\title{Domain shift-robust object detection with GenAI image editing}

\author[a,b]{Isabel D. Stein}
\author[a]{Thijs A. Eker}
\author[a]{Sebastiaan P. Snel}
\author[a]{Ella P. Fokkinga}
\author[a]{Klamer Schutte}
\author[b]{Luca Ambrogioni}
\author[a]{Friso G. Heslinga}
\affil[a]{TNO - Defence, Security and Safety, the Hague, the Netherlands}
\affil[b]{Radboud University, Nijmegen, the Netherlands}
\authorinfo{Corresponding author: Friso G. Heslinga. E-mail: fgheslinga@gmail.com}

\begin{document} 
\maketitle

\begin{abstract}

Object detectors often degrade under domain shifts such as changes in lighting, weather, or occlusion. These shifts alter object appearance and expose a reliance on visual shortcuts learned from the training distribution that do not generalize across domains. Acquiring sufficient real-world samples to capture such domain variation is particularly difficult in specialized, low-data settings. Recent advances in diffusion-based generative image editing have shown promise for improving the in-domain performance of object detectors through synthetic data augmentation. However, their potential to improve out-of-domain robustness remains largely unexplored. We hypothesize that generative image editing can simulate a controlled domain shift in training data, effectively bridging the gap between source and target domains. 

To test this, we studied camouflaged military vehicle detection as a challenging domain shift scenario. Detectors trained on uncamouflaged data demonstrate substantial degradation on real test imagery containing foliage, netting, and multi-spectral camouflage across 15 vehicle classes in close-up, ground-level imagery. We used two diffusion-based editing models, Qwen Image Edit 2509 and Flux.2 Dev, to synthetically add camouflage to the training data, alongside a LoRA fine-tuned version of Qwen. A non-generative black-bar occlusion baseline served as a lower bound on augmentation quality. Using a GroundingDINO detector trained on real and synthetic data, generative camouflage augmentation yielded substantial mAP improvements for foliage ($+20.1$) and netting ($+14.4$) camouflage. Generating multi-spectral camouflage proved more challenging, but LoRA fine-tuning improved performance by 4.4 mAP over the uncamouflaged baseline. The stronger gains from more realistic augmentations relative to the black-bar baseline confirm that higher synthetic fidelity introduces meaningful visual features. These findings demonstrate that generative image editing is a promising approach to reducing domain gaps in low-data regimes, with fine-tuning available to address cases where zero-shot editing alone falls short.

\end{abstract}

\keywords{Diffusion model; Rectified flow transformer; Data augmentation; Generative AI; Synthetic data; Deep learning; Object detection; Vehicle detection; Domain shift; Camouflage}

\section{Introduction}
\label{sec:intro} 

Object detection models often  degrade substantially when deployed in an environment that differs from their training data, \cite{Ben-David2010-bq} a phenomenon known as domain shift. Domain shifts can arise from changes in environmental conditions, object appearance, or background context, which alter the visual cues available for detection. \cite{Hendrycks2019-rf} Detectors are particularly vulnerable to these changes when they rely on visual shortcuts that are predictive in the training domain but fail to generalize across domains. \cite{Geirhos2020-dl} Robust detection therefore relies on learning feature representations that remain informative despite domain shifts. \cite{Hofmeijer2026} However, capturing the full range of target-domain variations in real-world data is often infeasible. \cite{wang2022generalizing} 

Recent advances in generative AI (GenAI), particularly diffusion-based image synthesis, offer a promising approach to enrich limited real-world datasets with synthetic diversity. \cite{fokkinga2025generative} Diffusion-generated images have been shown to improve in-domain segmentation, classification, and detection performance \cite{tan2025generative,azizi2023synthetic,Hu2025-hn,Wolk2026-xu,Fokkinga2026-or,snel2025data,Eker2026}, yet to our knowledge this approach has not been used to improve out-of-domain detection by editing source-domain images, without target-domain data. In this work, we hypothesize that generative image editing can reduce the source-target domain gap by introducing controlled target-domain appearance variations into the training data while preserving the underlying object and scene content.

We focus on camouflaged military vehicle detection as a challenging domain shift. Detectors trained on uncamouflaged vehicles are expected to experience substantial performance degradation when deployed on camouflaged vehicles, partly due to the occlusion of key visual cues and partly because camouflage reduces the contrast between the target and the background. At the same time, collecting and annotating large amounts of real-world camouflaged military imagery is particularly difficult. This scarcity motivates the use of generative image editing to construct a synthetic camouflaged dataset. A dataset containing imagery of 15 uncamouflaged military vehicle classes is used, onto which synthetic camouflage is introduced while preserving the underlying vehicle geometry. This introduces target-domain appearance variation without requiring additional data collection. Generative image editing is particularly promising in this setting because modern generative models can synthesize images with high fidelity \cite{rombach2021highresolution, Esser2024-ro}, reproducing detailed visual cues such as texture, shadow, and reflections, which support fine-grained recognition.

In this work, we evaluate whether generative image editing can reduce detection performance degradation under camouflage-induced domain shifts in low-data settings. We further investigate whether fine-tuning generative models on domain-specific data improves the usefulness of the resulting synthetic data for object detection. To address these questions, we compare zero-shot and fine-tuned generative image editing pipelines that synthetically introduce foliage, netting, and multi-spectral camouflage. A non-generative augmentation baseline is included to assess the importance of synthetic realism.

\section{Related works}
\label{sec:relatedworks}

Data augmentation helps models learn domain-invariant features that remain consistent across various transformations. \cite{Enomoto2025-ad, Noori2024-yi} Through diverse augmentations models are encouraged to ignore source-specific biases, such as camera color characteristics, and instead focus on universal semantic cues. \cite{Wang2025-nv} This improves robustness to distribution shifts while mitigating overfitting by exposing models to a wider range of training scenarios.

\subsection{Classical augmentation strategies}
Information dropping augmentations improve robustness by deliberately removing parts of the image to simulate occlusion, conceptually related to camouflage, where physical concealment obscures parts of the object's recognizable structure rather than removing pixels entirely. Cutout randomly masks square regions with a constant value. \cite{DeVries2017-zi} Random Erasing extends this idea with variable replacement values. \cite{Zhong2017-ct} GridMask uses a structured grid to balance information deletion and preservation. \cite{Chen2020-wm} Hide-and-Seek divides the image into patches and hides them with a given probability, encouraging the model to identify objects from their full spatial extent. \cite{Singh2017-bu} Progressive masking approaches such as Granular Ball Guided Masking (GBGM) further refine this idea by adaptively preserving semantically important regions while masking less informative areas. \cite{Xia2026-rv} Camouflage can be seen as a natural, real-world instance of information dropping, where texture and pattern reduce the visual discriminability of an object rather than removing pixels entirely.

While previous techniques modify or remove information within a single image, more recent approaches increase diversity by combining multiple training samples, helping reduce spurious correlations between objects and their typical backgrounds. \cite{Zeng2024-jx, Wang2024-gy} MixUp generates new samples by linearly interpolating pairs of images and labels, encouraging smoother decision boundaries and improved generalization. \cite{Zhang2017-hi} LossMix extends this idea by interpolating prediction losses instead of labels, enabling learning from partially visible objects in mixed images. \cite{Vu2023-jd}. CutMix replaces image regions with patches from other images while mixing labels proportionally to the replaced area, encouraging recognition from partial features. \cite{Yun2019-iy} Mosaic augmentation combines four images into a single grid to expose the model to multiple contexts simultaneously, benefiting small object detection. \cite{Bochkovskiy2020-me} Copy-Paste augmentation further diversifies training data by inserting object instances into new backgrounds to break spurious object–context associations. \cite{Ghiasi2020-kk}

\subsection{Synthetic dataset augmentation}
In specialized low-data domains, such as military vehicle detection, real-world datasets are often small, sensitive, and difficult to expand. 3D-model based simulations have shown to be an effective source of training data \cite{Eker2023, Heslinga2024combining, Heslinga2024Simulation}, but mostly to increase in-domain variation, not to augment existing image data. Classical augmentation is fundamentally constrained to the visual information present in the training set: geometric and photometric transformations can only recombine existing pixel content and cannot introduce new semantic content or stylistic diversity. \cite{Shorten2019-se, Trabucco2023-xs} 
This motivates the use of generative models, which overcome this limitation by synthesising semantically novel imagery.

Modern generative models have made high-fidelity, text-conditioned image synthesis available \cite{rombach2021highresolution, Esser2024-ro}, with demonstrated improvements in downstream object detection performance across low-data and domain-specific settings. \cite{tan2025generative,Wolk2026-xu,fokkinga2025generative}
Approaches for generative augmentation in detection broadly fall into two categories: text-to-image (T2I) generation, which synthesises new training images from prompts, and image-to-image (I2I) editing, which modifies existing images either globally through instruction-based editing or locally through masked inpainting, each offering different trade-offs between diversity, spatial consistency, and annotation preservation.

Recent studies across medical imaging, agriculture, logistics, and remote sensing demonstrate the effectiveness of diffusion-based augmentation over traditional geometric transformations. In medical segmentation, AugPaint preserves foreground annotations while synthesizing novel backgrounds through latent diffusion inpainting, maintaining exact alignment between synthetic images and label masks. \cite{Hu2025-hn} Similar benefits have been observed across diverse low-data domains \cite{tan2025generative}, where diffusion-based augmentation introduces greater semantic and environmental diversity than classical transformations, improving robustness under varying conditions. Collectively, these studies suggest that diffusion-based augmentation can improve robustness by introducing environmental diversity that classical transformations cannot, while preserving task-relevant spatial structure.

\subsection{Domain-targeted augmentation and adaptation}
Off-the-shelf diffusion models are trained on large-scale, general-domain datasets and may not capture domain-specific characteristics, leading to inconsistencies in object geometry, scale, and viewpoint that limit their utility for specialized applications.\cite{Fokkinga2026-or, Zhu2024-ls} In low-data domains such as camouflaged military vehicle detection, where real-world datasets are small and difficult to expand, fine-tuning of a generative model might be needed. Low-Rank Adaptation (LoRA) enables this efficiently: by injecting trainable low-rank updates into selected model layers while keeping the majority of weights fixed, a generative model can be adapted to a specific domain from a small set of examples without full retraining.\cite{hu2021lora} This is especially practical when a small amount of images is used to train the detector can also serve as fine-tuning data for the generative model. Fokkinga et al.\ \cite{Fokkinga2026-or} demonstrate this directly in the military domain, fine-tuning FLUX on as few as 8 real vehicle images per class using LoRA and achieving meaningful detection improvements, confirming that domain-specific adaptation is feasible even under severe data constraints. Recent work further confirms that aligning synthetic data with the target domain improves detection performance, and that generative augmentation can transfer more reliably than photorealistic 3D simulation in low-data settings.\cite{tan2025generative, Wolk2026-xu}
 
Most closely related to our work, DODA proposes a diffusion-based framework for adapting object detectors to new domains by generating scene layout-conditioned detection data guided by domain embeddings extracted from unlabeled target-domain reference images. \cite{Xiang2024-vi} While DODA demonstrates that diffusion-generated data can bridge domain gaps, it relies on layout-to-image generation and requires unlabeled reference images from the target domain to guide generation. More broadly, existing work on generative augmentation for domain adaptation focuses on naturally occurring distribution shifts such as cross-farm agricultural variation targeted by DODA and indoor-to-outdoor agricultural conditions \cite{tan2025generative}. In contrast, our approach targets physical concealment as a domain shift, where camouflage is intentionally applied to reduce object detectability, and uses image-to-image editing applied directly to source-domain images. Our zero-shot pipeline requires no target-domain data whatsoever, while our LoRA pipeline uses real camouflaged images only to fine-tune the generative model, never to train the detector. We investigate whether this approach can bridge the domain gap across three structurally distinct camouflage types, evaluating both a zero-shot pipeline and a LoRA fine-tuned pipeline.

\section{Methods}
\label{sec:methods}
\begin{figure}
    \centering
    \includegraphics[width=1\textwidth]{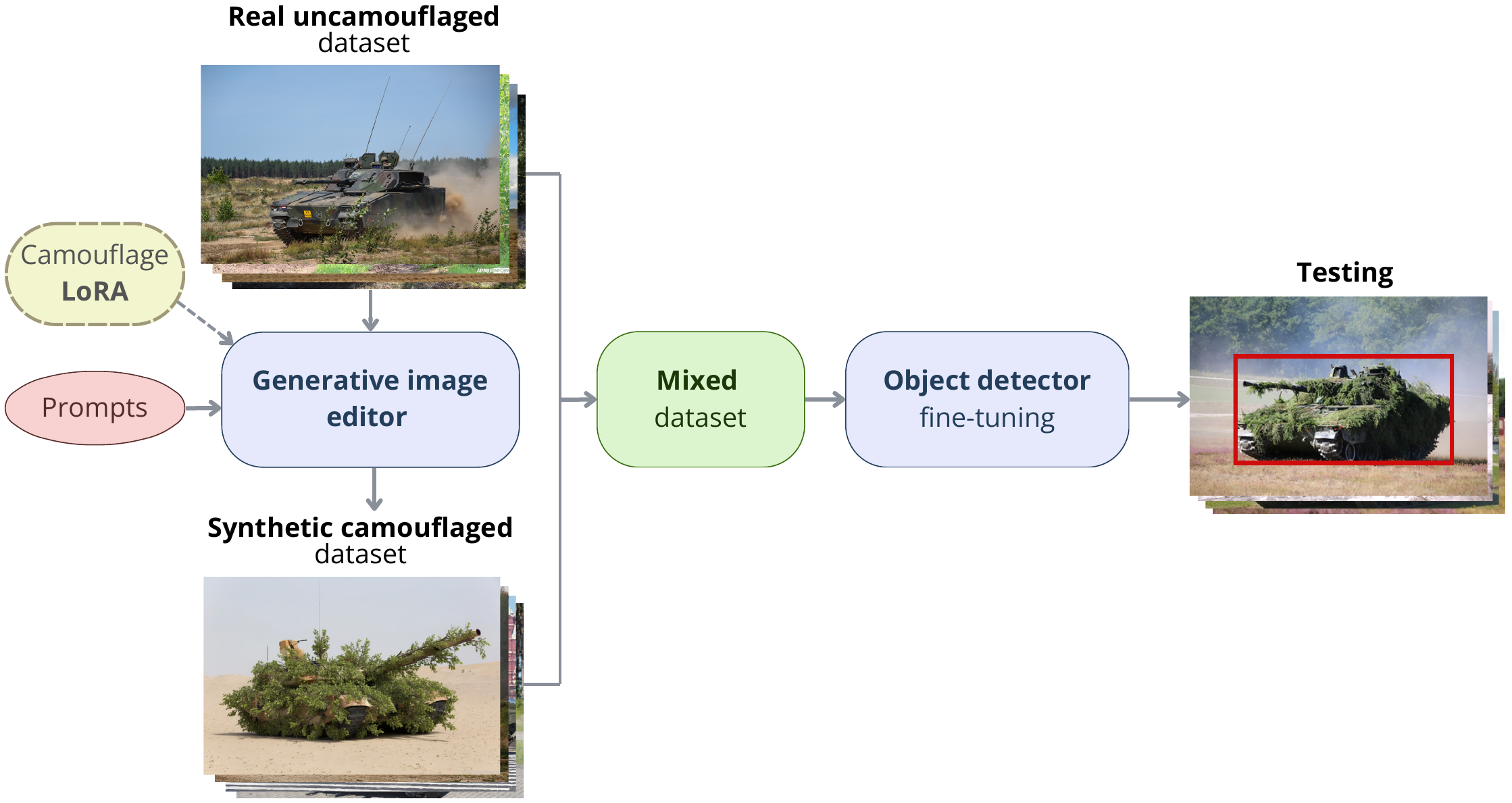}
    \vspace{2pt}
    \caption{Overview of our synthetic camouflage augmentation pipeline. Generative image editing is used to introduce synthetic camouflage into uncamouflaged vehicle images based on text prompts, optionally supported by LoRA fine-tuning to improve generation quality. For LoRA finetuning, a separate set of camouflaged is used, not shown in this figure. The synthetic and real images are combined in a mixed dataset to fine-tune a GroundingDINO detector, which is then evaluated on real-world camouflaged vehicles.}
    \label{fig:pipeline}
\end{figure}

The experimental setup is designed to assess whether synthetic camouflage augmentation can reduce the domain gap between uncamouflaged and camouflaged vehicle detection. To this end, multiple synthetic augmentation strategies are used to construct training datasets, after which object detectors are trained and evaluated on real-world camouflaged vehicle imagery. The four evaluated approaches comprise two zero-shot generative image editing setups (Flux.2 Dev, Qwen Image Edit 2509), a LoRA fine-tuned variant (of Qwen Image Edit 2509), and a non-generative augmentation baseline. The generative pipeline consists of (i) prompt engineering per camouflage type and model, (ii) synthetic camouflage generation via zero-shot or LoRA fine-tuned editing, and (iii) downstream use in object detector fine-tuning. The effect of synthetic data quality is further investigated through quality filtering and post-processing procedures. An overview of the synthetic camouflage augmentation pipeline is illustrated in Figure~\ref{fig:pipeline}.

\subsection{Real dataset}
\label{methods:datasets:real}
The uncamouflaged real-world dataset consists of 15 military vehicle classes, with 360 training images (24 per class), 150 validation images (10 per class), and 432 test images. Camouflaged imagery is only available for seven vehicle classes (CV90, PzH 2000, M1A2, T-90, Leopard 2, M109, and Fennek). Consequently, camouflage evaluation is restricted to this subset. The corresponding uncamouflaged test subset contains 190 images and serves as the reference against which performance on the camouflaged test sets is compared. Six of these classes share comparable structural features such as barrel, turret, hull, and tracks, making this a fine-grained recognition task in which the detector must distinguish visually similar classes despite substantial intra-class appearance variation. 

\begin{figure}[H]
\centering
\newlength{\imgheight}
\setlength{\imgheight}{0.13\textheight}
\setlength{\tabcolsep}{1pt}
\begin{tabular}{cccc}
\textbf{Uncamouflaged} & \textbf{Foliage} & \textbf{Netting} & \textbf{Multi-spectral} \\
\includegraphics[height=\imgheight]{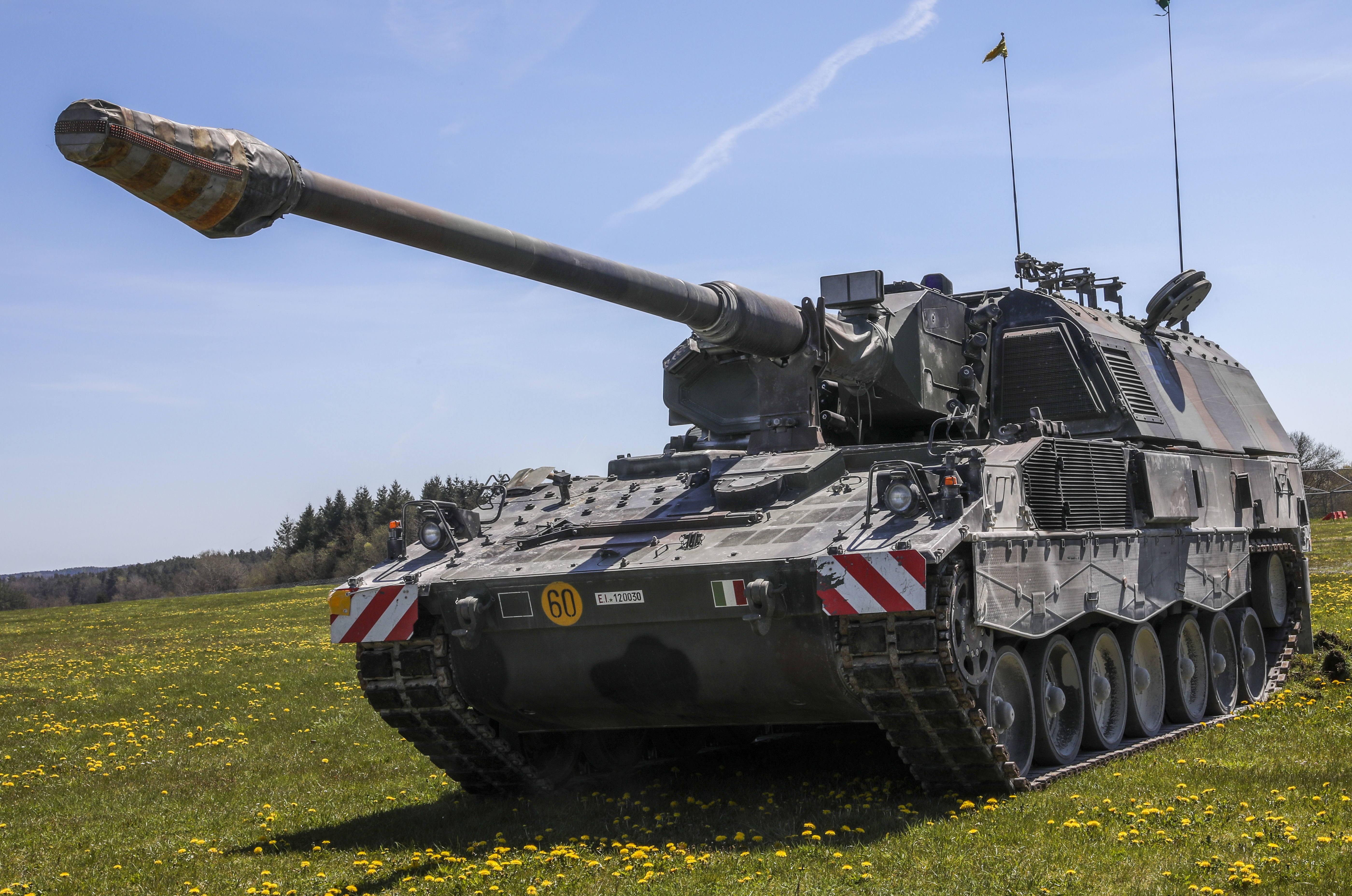} &
\reflectbox{\includegraphics[height=\imgheight]{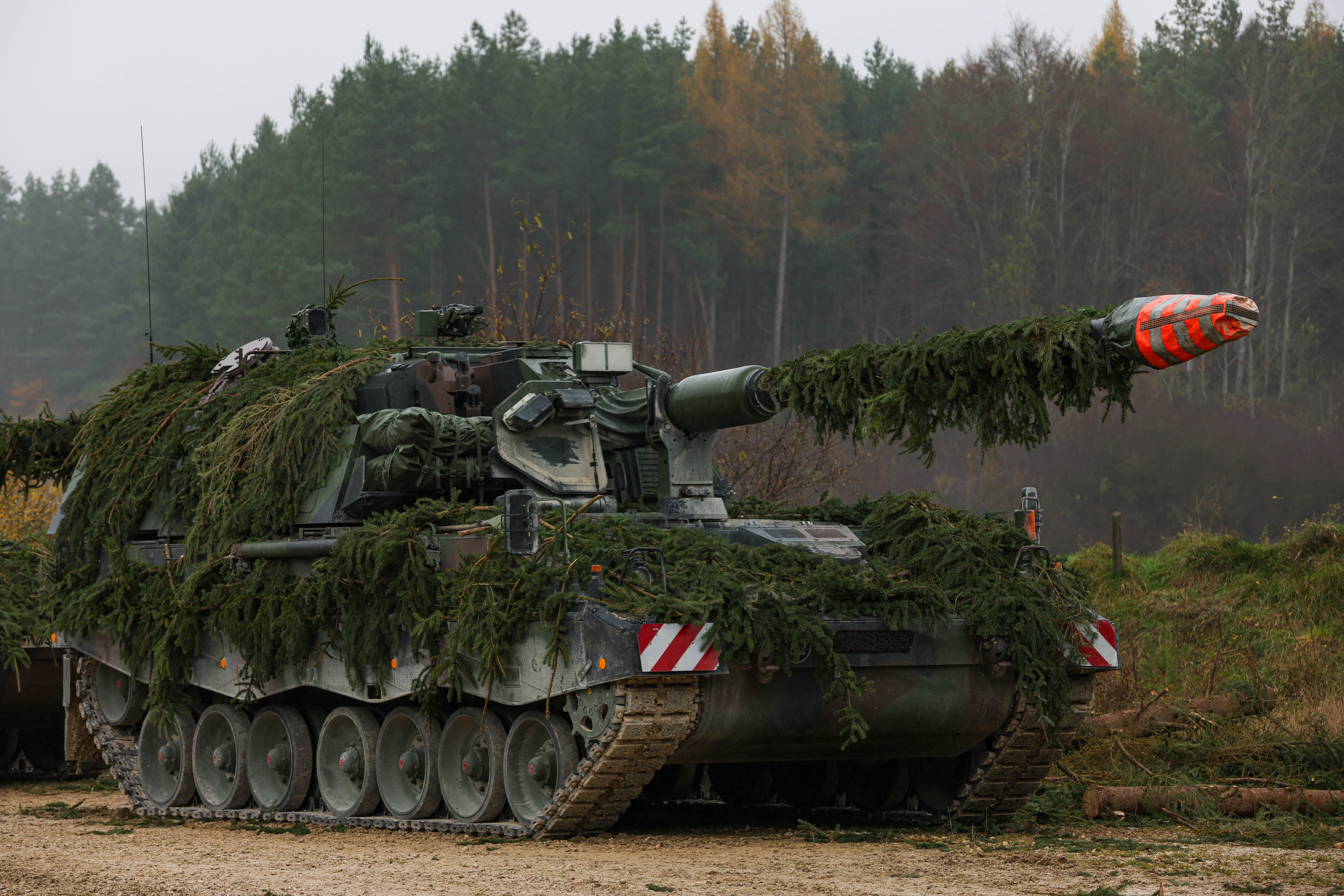}} &
\reflectbox{\includegraphics[height=\imgheight]{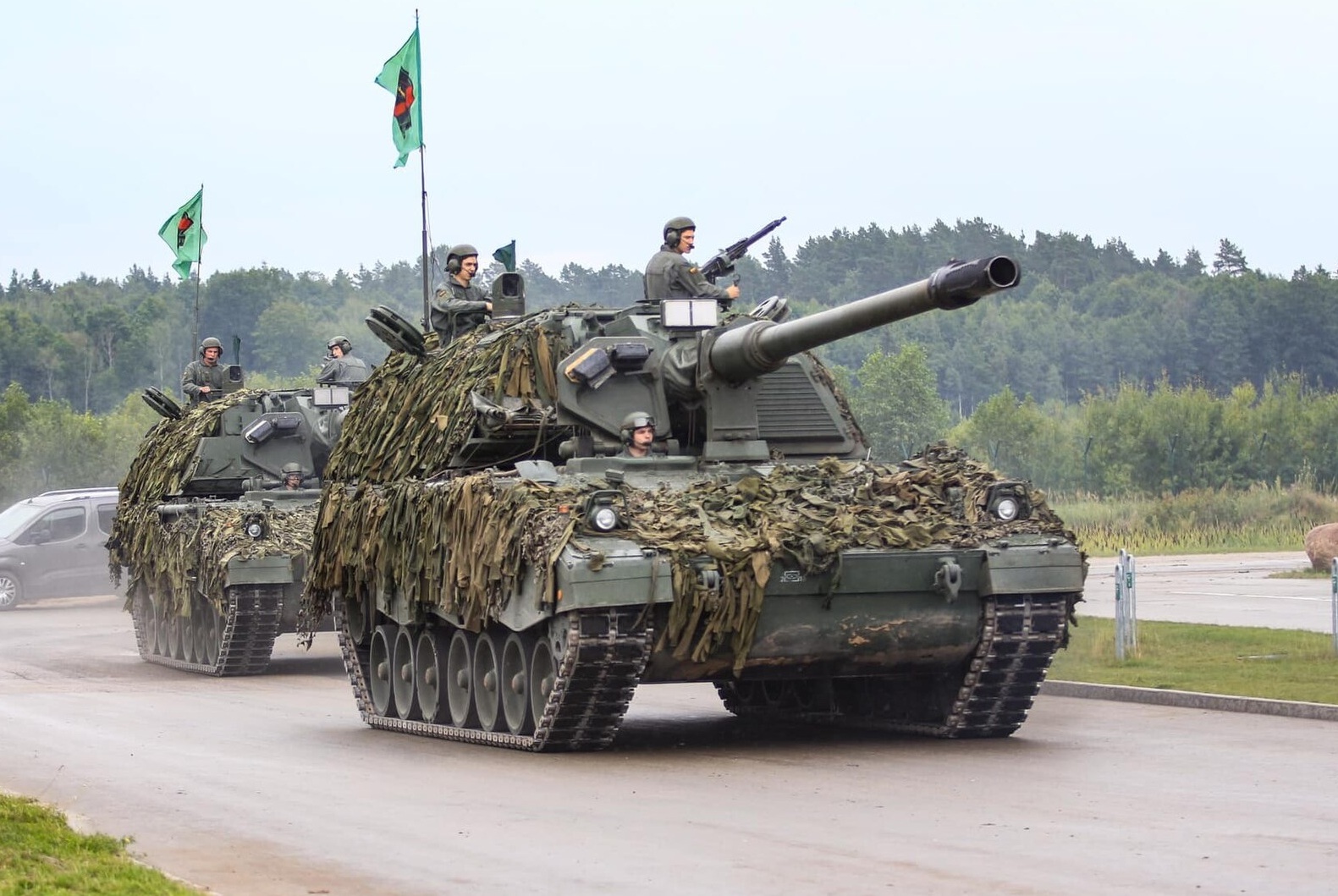}} &
\includegraphics[height=\imgheight]{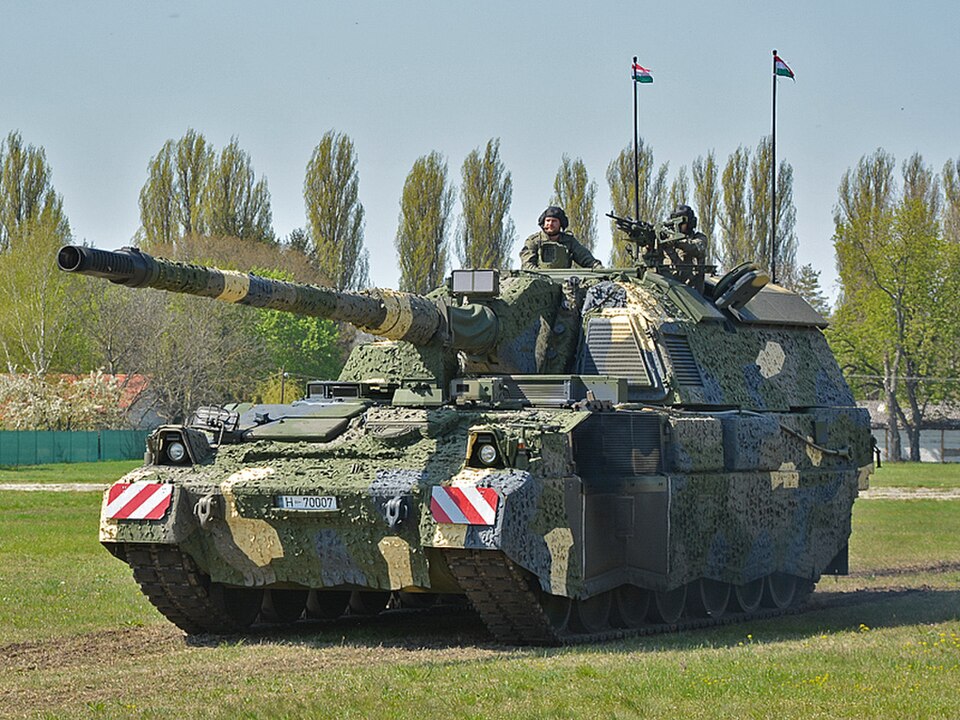} \\
\end{tabular}
\vspace{4pt}
\caption{Examples of the PzH 2000 self-propelled howitzer under different camouflage conditions, from left to right: uncamouflaged, foliage camouflage with branches, standard netting, and multi-spectral camouflage. Source images 1--2: DVIDS, public domain. Image 3: Wikimedia Commons, GNU FDL v1.2+. Image 4: Admiralis-generalis-Aladeen, Wikimedia Commons, CC BY-SA 4.0.}
\label{fig:test_subset}
\end{figure}

The camouflaged real-world test set was assembled from publicly available web-sourced imagery and covers three camouflage types: foliage, netting, and multi-spectral camouflage. Because the imagery is predominantly close-up and ground-level, the vehicles generally remain clearly visible under camouflage. Consequently, the primary challenge is not object localization but fine-grained recognition, where camouflage obscures the visual cues needed to distinguish between structurally similar vehicle classes. Figure~\ref{fig:test_subset} illustrates the appearance variation introduced by each camouflage type for the Panzerhaubitze 2000 class. Foliage and netting camouflage are evaluated across all seven classes, comprising 93 and 141 images, respectively. Multi-spectral camouflage is evaluated on four classes only (Fennek, M1A2, CV90, and Leopard), comprising 69 images, as camouflaged imagery for the remaining classes could not be sourced. In total, the camouflaged test set contains 303 images.

\subsection{Synthetic camouflage generation}
This work evaluates two open-weight generative image editing models: Qwen Image Edit 2509 (hereafter referred to as \textit{Qwen}) and Flux.2 Dev (hereafter referred to as \textit{Flux}). Both are state-of-the-art diffusion-based image generation models capable of editing existing images based on textual instructions. Qwen is built upon a multimodal diffusion transformer architecture designed for joint reasoning over text and image content, enabling accurate instruction-following while preserving image identity during editing. \cite{Wu2025-bb} Flux is a more general-purpose flow-matching image generation and editing model optimized for high-quality image synthesis. \cite{Black_Forest_Labs2025-ci}

\subsubsection{Zero-shot image editing}
\label{methods:augmentation_pipelines:generation}
Prompts were designed to introduce foliage, netting, and multi-spectral camouflage while preserving vehicle structure and scene context. To enable a fair comparison between editing models, prompts were aligned as closely as possible in terms of camouflage type, placement, and coverage. Because multi-spectral camouflage is poorly represented in general-domain training data, it was specified through descriptive visual characteristics rather than a known category label. Foliage and netting camouflage are generated as mixtures of multiple subtypes. Foliage comprises branches, hay, and moss, while netting comprises standard netting, tarp, and hessian variants, combined in a 16:4:4 ratio.

Synthetic image generation for both models was implemented using ComfyUI \cite{comfyui_2023} with seeds randomized per image to maximize output diversity. Qwen Image Edit 2509 was loaded in FP8 quantization with the Qwen 2.5 VL 7B text encoder, also quantized to FP8. Sampling used the Euler scheduler with 50 steps, a guidance scale of 4, a sampling shift of 3, and a denoise strength of 1.0. When a LoRA adapter (Section \ref{methods:LoRA}) was applied, the strength was set to 1.0 for foliage and netting. For multi-spectral camouflage, a higher strength of 1.15 was used, which produced coverage patterns more consistent with the real multi-spectral training imagery, including partial obscuring of wheels and barrel areas that is characteristic of this camouflage type. FLUX.2 Dev was loaded in FP8 mixed quantization with a Mistral 3 Small text encoder in bf16 precision. Sampling used the Euler scheduler with 30 steps and a guidance scale of 4. All generation was performed on a NVIDIA L4 (22.5 GB VRAM).

\subsubsection{LoRA fine-tuning}
\label{methods:LoRA}
Zero-shot editing relies entirely on the generative model's pretrained priors, which may not capture the specific visual characteristics of each camouflage type. To investigate whether fine-tuning the model to real camouflaged examples improves synthesis quality and downstream detection performance, LoRA fine-tuning \cite{hu2021lora} is applied to Qwen across all three camouflage types. LoRA adapts a pretrained model by introducing a small number of trainable low-rank parameters while keeping the original model weights fixed.

\begin{figure}
    \centering
    \begin{tabular}{c@{\hspace{4pt}}ccc}
        \rotatebox{90}{\parbox{3cm}{\centering Real camouflaged (after)}} &
        \includegraphics[height=3cm]{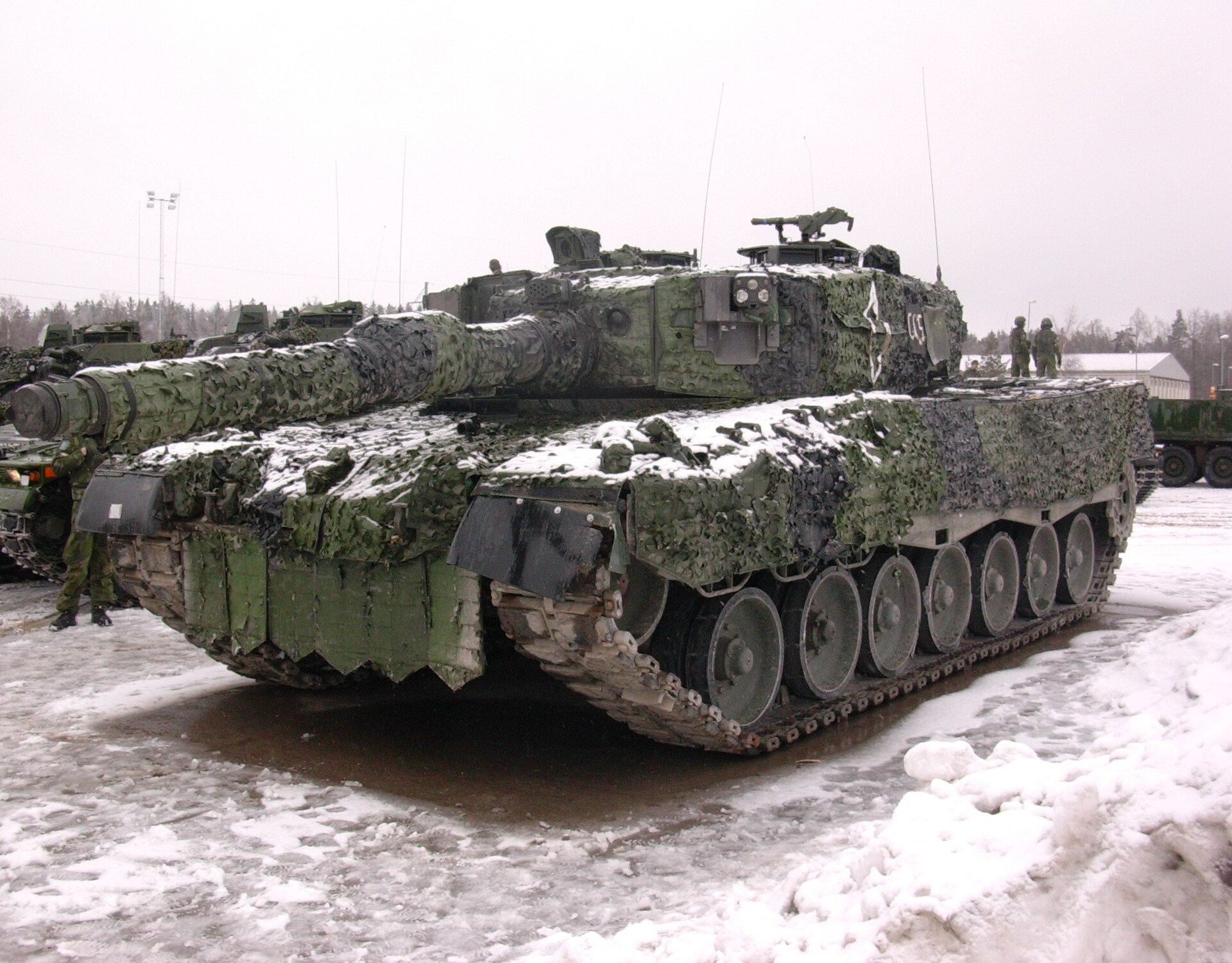} &
        \includegraphics[height=3cm]{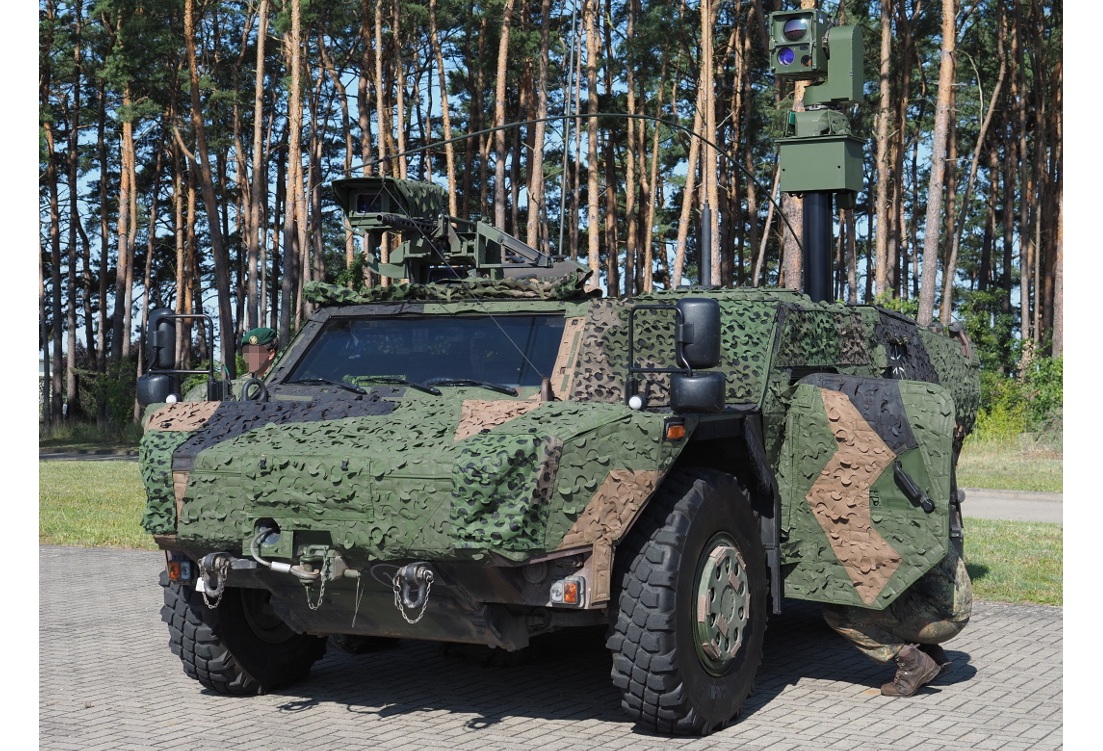} &
        \includegraphics[height=3cm]{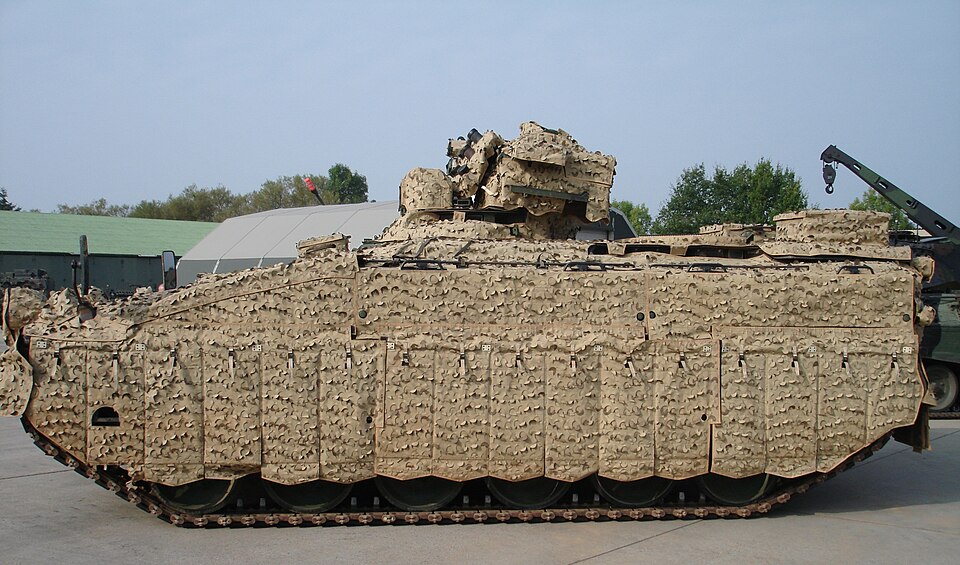} \\[4pt]
        \rotatebox{90}{\parbox{3cm}{\centering Synthetic uncamouflaged (before)}} &
        \includegraphics[height=3cm]{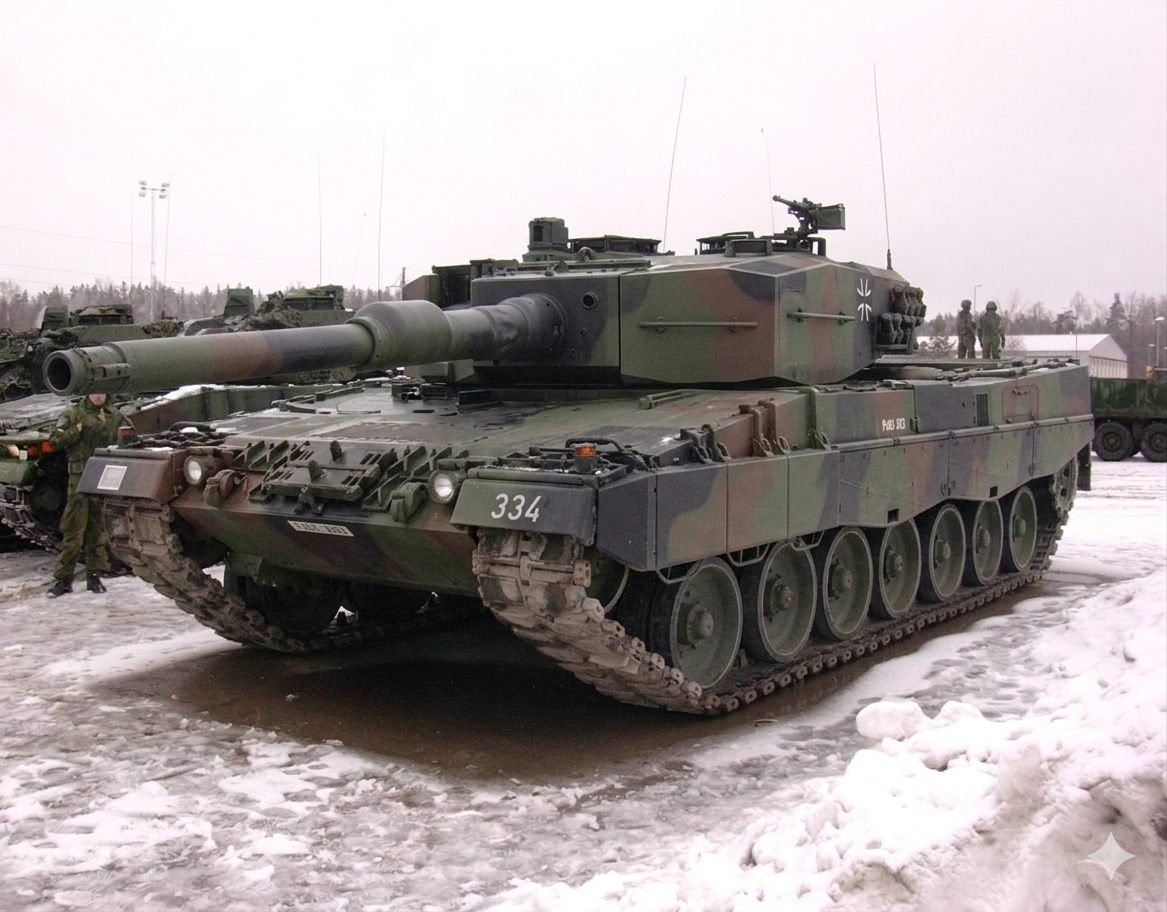} &
        \includegraphics[height=3cm]{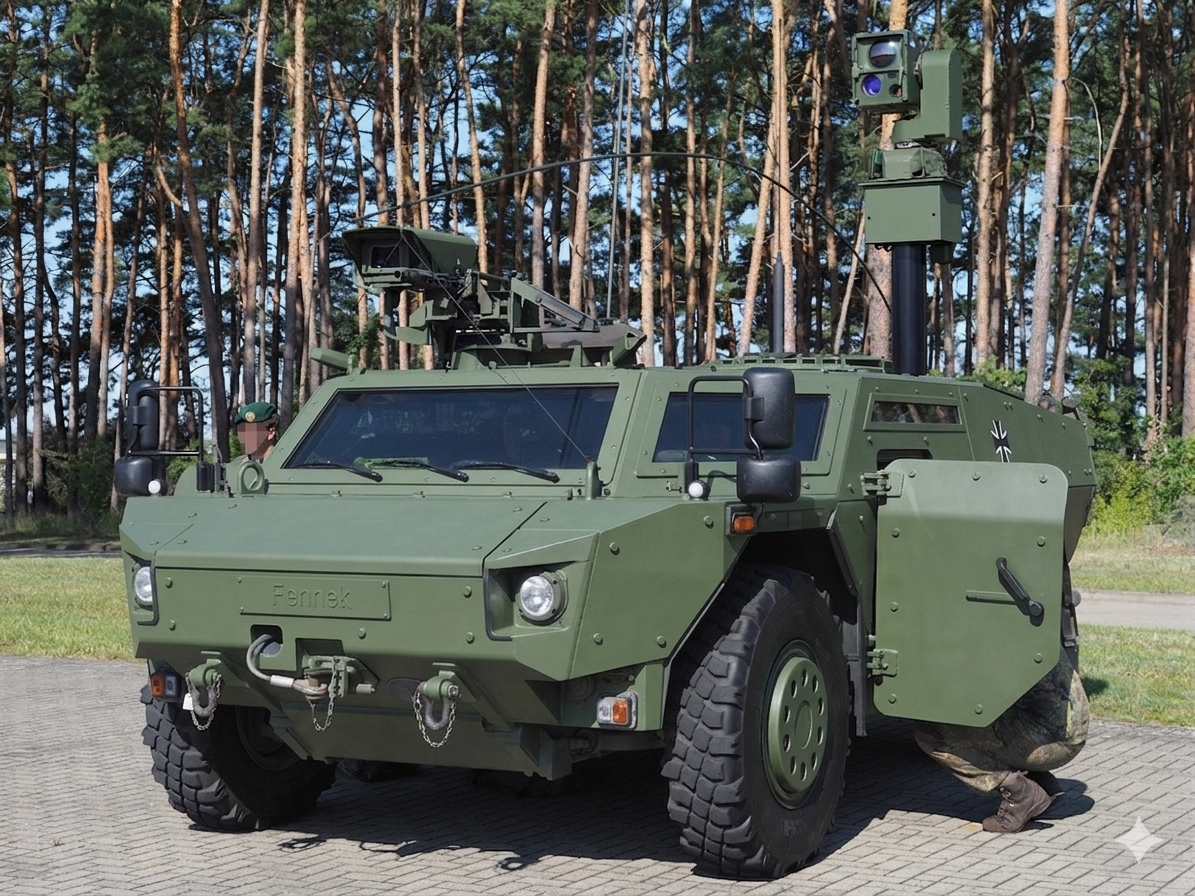} &
        \includegraphics[height=3cm]{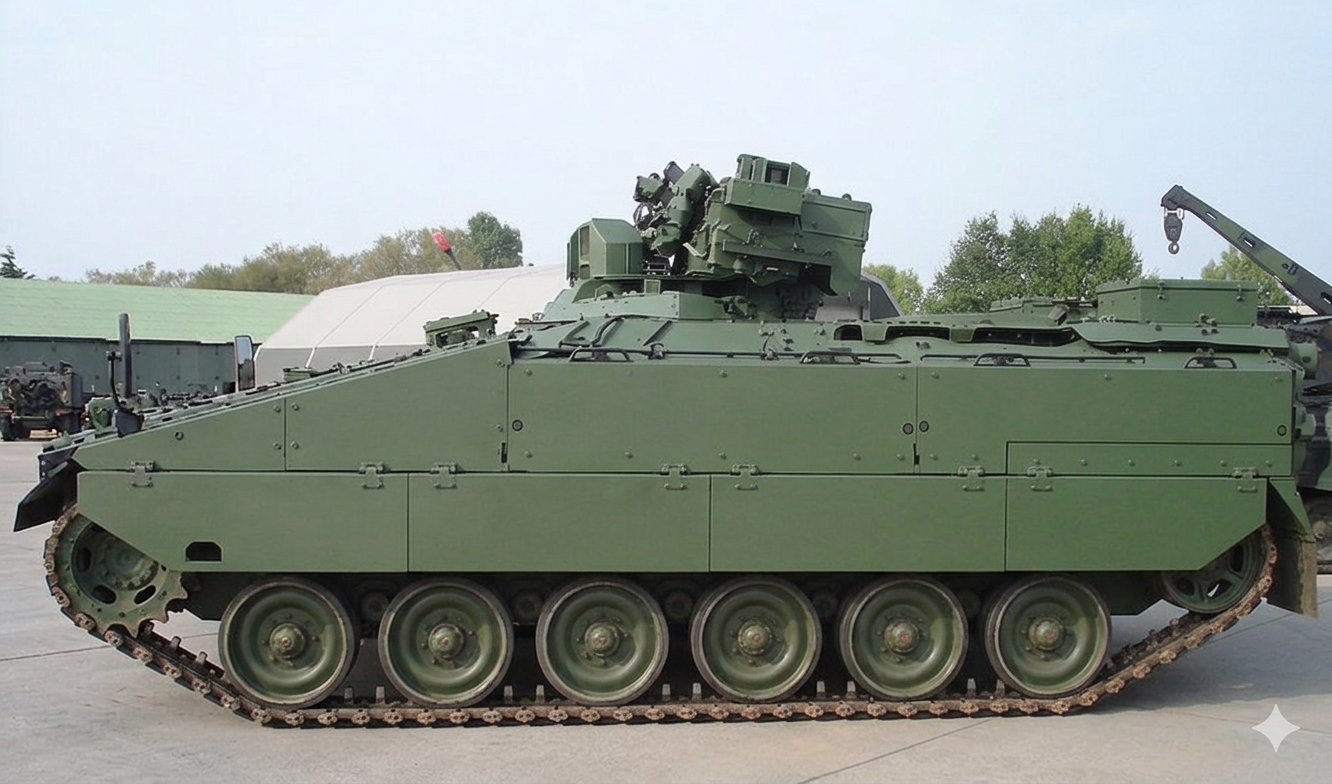} \\
    \end{tabular}
    \caption{Training pairs for LoRA fine-tuning. Each column shows a matched real (camouflaged) and  synthetic (uncamouflaged) image pair. From left to right: Boevaya mashina, CC BY-SA 3.0 DE; Mr Bullitt, CC BY-SA 2.5; Sonaz, CC BY 3.0. All via Wikimedia Commons.}
    \label{fig:lora-pairs}
\end{figure}

A separate LoRA adapter is trained for each camouflage type using paired before/after images, as illustrated in Figure~\ref{fig:lora-pairs}. The fine-tuning dataset comprises 37 image pairs for foliage, 24 for netting, and 14 for multi-spectral camouflage. For foliage and netting, fine-tuning is applied only to the dominant subtype of each camouflage category (branches and standard netting, respectively), as too few real images were available to train adapters for the rarer subtypes. The images used for LoRA fine-tuning are distinct from those in the camouflaged test set, with no image-level overlap, and include both evaluation vehicle types and other vehicle types. Since paired real images of the same vehicle before and after camouflage application are unavailable, the required before/after pairs are constructed synthetically. Camouflage is first removed from real camouflaged vehicle images using Qwen, producing synthetic uncamouflaged counterparts that serve as the ‘before’ images. Because only a limited number of pairs is required and this stage is not time-constrained, multiple candidate edits are generated for each image and the most plausible result is selected manually. The resulting image pairs are verified through visual inspection before being used for fine-tuning.

For foliage and netting camouflage, adapters were trained on 37 and 24 image pairs respectively, using a rank of 32, linear alpha of 16, and a learning rate of $5\times10^{-5}$. For multi-spectral camouflage, a higher-capacity adapter was trained on 14 image pairs using a rank of 64, linear alpha of 32, and a learning rate of $1\times10^{-4}$. Based on visual inspection, training was continued for 1950 steps for netting and 3000 for both foliage and multi-spectral. All adapters were trained with a batch size of 1, gradient accumulation of 1, and caption dropout set to 0.15. The text encoder was kept frozen throughout; only the U-Net was updated, with gradient checkpointing enabled to reduce VRAM usage. Training used the AdamW 8-bit optimizer with a flowmatch noise scheduler and bf16 precision. The base model was quantized to uint3 with accuracy recovery adapters, and the text encoder to qfloat8, to fit within the available hardware budget. Input images were processed at resolutions of 512, 768, and 1024 pixels.

Each image pair is captioned with a minimal before/after description: the before caption labels the image as ‘a vehicle’, and the after caption describes the camouflage appearance. For foliage, the after caption varied coverage level across light, moderate, and heavy. For netting, both coverage level and color were varied. For multi-spectral, only color was varied. Image pairs were collected across a variety of vehicle types, including types not present in the camouflaged test set, to encourage generalization of the learned camouflage appearance across different vehicle shapes. All LoRA training was performed on a single NVIDIA A40 (45 GB VRAM).

\subsubsection{Annotation of synthetic images}
Generated images were automatically annotated with vehicle bounding boxes using GroundingDINO in zero-shot, open-vocabulary mode. GroundingDINO was prompted with a class-specific text prompt listing all 15 vehicle classes, retaining all detections above a confidence threshold of 0.35 and a minimum area ratio of 0.05 relative to the image area. Class labels were inherited from the corresponding source images, ensuring label consistency across the generated dataset. Visual inspection was performed to correct false or missing detections.

\subsubsection{Quality filtering and post-processing}
\label{methods:postprocessing}
\begin{figure}[H]
    \centering
    \begin{tabular}{ccc}
        \includegraphics[width=0.28\textwidth]{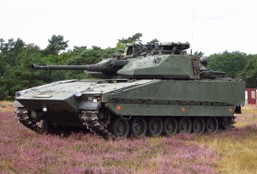} &
        \includegraphics[width=0.28\textwidth]{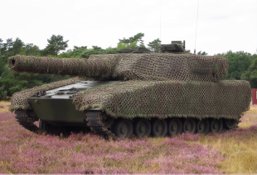} &
        \includegraphics[width=0.28\textwidth]{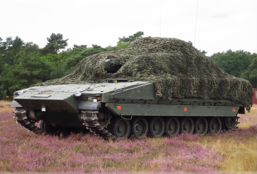} \\
        \footnotesize Original &
        \parbox{0.28\textwidth}{\centering\footnotesize Structural deformation \\ (unrealistically enlarged barrel)} &
        \parbox{0.28\textwidth}{\centering\footnotesize Incomplete rendering \\ (barrel removed)} 
    \end{tabular}
    \vspace{4pt}
    \caption{Examples of generation failures during quality filtering. The original uncamouflaged vehicle (left, Photo: Jorchr, CC BY-SA 3.0, via Wikimedia Commons) is shown alongside two faulty outputs produced when prompting for camouflage netting addition: structural deformation where the barrel is hallucinated as enlarged (center), and incomplete rendering where the barrel is removed entirely (right).}
    \label{fig:faulty_images}
\end{figure}

Generative models often introduce visual artifacts in specific parts of generated or edited images. These artifacts can cause the object detector to learn spurious features that do not generalize to real-world data, potentially degrading detection performance. To avoid this, we remove images from the dataset if they exhibit: hallucinated additions (new vehicles or guessed occluded parts); structural deformation (significantly altered silhouette); incomplete rendering; asset fusion (merged vehicles or objects); or physically impossible camouflage placement. Figure~\ref{fig:faulty_images} illustrates examples of structural deformation and incomplete rendering as observed during filtering. Detectors are trained on both the unfiltered and filtered datasets to quantify the impact of this step.

Two post-processing variants are evaluated to address artifacts introduced by the editing process. The first applies Gaussian blurring to reduce sharpness discrepancies between edited and unedited image regions. The second applies a diffusion-based refinement pass using Flux.1 Dev to improve visual consistency across the image. Both methods are intended to reduce generation artifacts that may lead the detector to learn editing-specific features rather than camouflage-related features.

\subsection{Non-generative baseline: black-bar augmentation}
As a lower bound on augmentation quality, a non-generative baseline is included to isolate the contribution of visual realism. This is done using an information-dropping strategy where black bars are overlaid onto the original images to simulate occlusion, with bar widths of 15, 20, 25, and 30 pixels and image-area coverage of 20, 25, 30, 35, or 40 percent. Each image is randomly assigned one width and coverage combination. The number of bars is determined by the target coverage fraction and bar width, ensuring the specified percentage of the image area is occluded. Bars are evenly spaced and rotated at a random angle before being cropped to the original image dimensions. Like generative camouflage augmentation, this approach partially occludes the vehicle, encouraging the detector to identify objects from partial visibility. However, unlike generative editing, the occlusion is artificial and uniform, providing no realistic camouflage appearance. This allows the contribution of visual realism to be isolated from the effect of partial occlusion alone.

\subsection{Experiments}
MM GroundingDINO detectors \cite{Zhao2024-oo} with a Swin-T backbone \cite{xie2021self} are initialized with weights pre-trained on Objects365, GoldG, GRIT-9M, and V3Det. A baseline detector is fine-tuned on 360 uncamouflaged vehicle images, establishing the performance degradation caused by camouflage-induced domain shift. The detector is then evaluated on the camouflaged test set described in Section~\ref{methods:datasets:real} using mean average precision across IoU thresholds from 0.50 to 0.95 (mAP) \cite{Padilla2020}, which combines classification accuracy with localization accuracy. We also report mAP scores for foliage, netting and multi-spectral subsets of the test set.

We experiment with each of the four generation approaches at varying proportions of original and synthetically camouflaged data (0--100\% synthetic in 25\% increments). The synthetic-to-real ratio is controlled by fixing the number of synthetic images at 1080 and oversampling the real training images (360 images) to achieve the target ratio. In addition, a combined augmentation experiment is conducted to assess whether the benefits of generative and non-generative augmentation are complementary. The synthetic portion is divided equally between Qwen-generated camouflage and black-bar augmentation.

Standard augmentations are applied during training, including photometric distortion, random horizontal flipping, minimum-IoU random cropping, and multi-scale resizing. All configurations are trained for approximately 64,800 iterations in total, with the number of epochs scaled inversely to dataset size to ensure a comparable computational budget across experiments. The best-performing checkpoint is selected based on validation mAP monitored every epoch. Optimization follows a cosine annealing learning rate schedule decaying from $1\times10^{-4}$ to $1\times10^{-5}$ over the first two-thirds of training, followed by a tenfold step decay. All experiments are repeated three times with different random seeds, and mean $\pm$ standard deviation is reported throughout.

\section{Results}

\begin{figure}
\centering
\includegraphics[width=\linewidth]{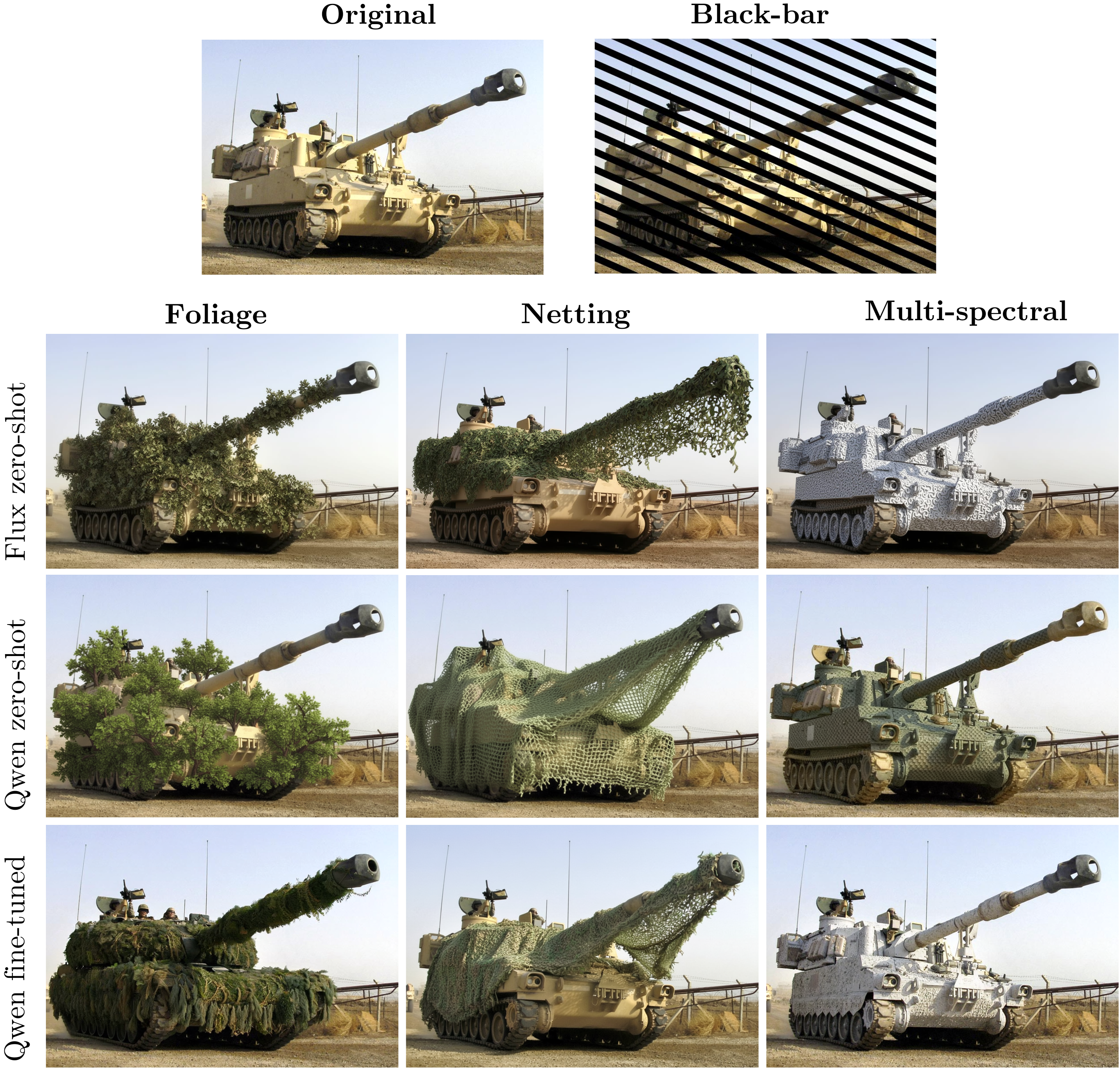}
\caption{Representative synthetic camouflage examples generated by each approach across all three camouflage types. The top row shows the original uncamouflaged vehicle for reference (Photo: Senior Airman Steve Czyz, U.S.\ Air Force, public domain, via Wikimedia Commons) as well as the black-bar augmentation.}
\label{fig:synthetic_examples}
\end{figure}

Figure~\ref{fig:synthetic_examples} shows a representative sample from each generative approach across all three camouflage types. Overall, zero-shot editing yielded visually plausible foliage and netting camouflage, whereas multi-spectral camouflage appeared less realistic under zero-shot editing and became more convincing only after LoRA fine-tuning.

\subsection{Baseline performance}
Table \ref{tab:performance} reports object detection performance across all training configurations and test conditions. The baseline detector, trained exclusively on uncamouflaged vehicles, exhibits a substantial performance drop when evaluated on camouflaged vehicles, falling from 86.6 to 56.8 mAP. Performance varies across camouflage types, with foliage and netting resulting in the largest degradation, while multi-spectral camouflage produces a comparatively smaller decrease.

\begin{table}
\centering
\caption{Object detection performance (mAP@50--95) on uncamouflaged and different camouflage test sets. Unless noted otherwise, models are trained on 75\% real (uncamouflaged) and 25\% synthetic (camouflaged) data. Values represent the mean $\pm$ standard deviation over three runs with different seeds. The approach combining Qwen \& black-bar augmentation was evaluated only on the overall uncamouflaged and camouflaged test sets.}
\vspace{3mm}
\label{tab:performance}
\begin{tabular}{llllll}
\toprule
Model & Uncamouflaged & Camouflaged & Foliage & Netting & Multi-spectral \\
\midrule \addlinespace[1mm] \textit{Baseline} \\ \cmidrule{1-1}
uncamouflaged & 86.6 $\pm$ 2.5 & 56.8 $\pm$ 1.8 & 54.1 $\pm$ 1.0 & 60.4 $\pm$ 3.3 & 75.4 $\pm$ 2.6 \\
\hline \addlinespace[1mm] \textit{FLUX} \\ \cmidrule{1-1}
zero-shot        & 87.3 $\pm$ 1.6 & 61.4 $\pm$ 5.2 & 59.1 $\pm$ 3.3 & 70.4 $\pm$ 4.7 & 66.5 $\pm$ 5.7 \\
\hline \addlinespace[1mm] \textit{Qwen} \\ \cmidrule{1-1}
zero-shot        & $\mathbf{87.7 \pm 2.3}$ & $\mathbf{69.9 \pm 2.1}$ & $\mathbf{74.2 \pm 1.6}$ & $\mathbf{74.8 \pm 0.8}$ & 70.9 $\pm$ 4.6 \\
+ finetuning  (25\% synthetic)    & 86.1 $\pm$ 3.0 & 67.5 $\pm$ 2.6 & 68.9 $\pm$ 1.4 & 74.3 $\pm$ 3.2 & 76.5 $\pm$ 6.4 \\
+ finetuning (50\% synthetic)    & 85.6 $\pm$ 3.2 & 64.9 $\pm$ 2.6 & 62.9 $\pm$ 3.8 & 72.3 $\pm$ 2.6 & $\mathbf{79.8 \pm 1.4}$ \\
\hline \addlinespace[1mm] \textit{Classical} \\ \cmidrule{1-1}
black-bar augmentation       & 87.5 $\pm$ 1.8 & 62.6 $\pm$ 1.8 & 58.6 $\pm$ 2.5 & 68.5 $\pm$ 4.2 & 71.2 $\pm$ 0.7 \\
black-bar augmentation \& Qwen  & 85.6 $\pm$ 2.9 & 69.9 $\pm$ 2.4 & --- & --- & --- \\
\hline
\end{tabular}
\end{table}

\subsection{Effect of synthetic augmentation}
Object detectors were trained across a range of synthetic data ratios (0--100\% in 25\% increments). The ratio of 25\% synthetic and 75\% real consistently yielded the best performance and is therefore the focus of the remaining results, with full results across all 
ratios provided in Section~\ref{supplementary:performance}. Table~\ref{tab:performance} reports detection performance for this configuration across all augmentation approaches, evaluated on both the uncamouflaged and camouflaged test sets including foliage, netting, and multi-spectral subsets. 

Qwen zero-shot achieves the highest overall camouflaged detection performance, reaching 69.9 mAP, an improvement of $+13.1$ mAP over the baseline, $+7.3$ mAP over the black-bar augmentation baseline, and $+8.5$ mAP over Flux zero-shot. A mixed training set combining real uncamouflaged images (75\%), Qwen zero-shot edited images (12.5\%), and black-bar augmented images (12.5\%) yields the same overall camouflaged mAP of 69.9. 

At the best-performing ratio of 25\% synthetic data, performance on the uncamouflaged test set remains stable across all augmentation approaches, ranging from 85.6 to 87.7 mAP. This confirms that synthetic camouflage augmentation does not degrade in-domain detection performance. To investigate whether the performance gap between Flux and Qwen was caused by sharpness discrepancies between Flux-generated and real images, Gaussian blurring and a second diffusion pass were evaluated as post-processing strategies; neither led to consistent improvement. Full results are provided in Section~\ref{supplementary:performance}.

Across all generative approaches, the largest improvements are observed for foliage and netting camouflage, while gains for multi-spectral camouflage are more limited. This is consistent with the baseline evaluation, where foliage and netting caused the largest performance drop and therefore offered the greatest room for improvement.

\subsubsection{Fine-tuned Qwen}
Fine-tuning Qwen with LoRA does not improve overall camouflaged performance (67.5 vs.\ 69.9 mAP), but produces consistent gains for multi-spectral camouflage, increasing from 70.9 to 76.5 mAP at 25\% synthetic data. Increasing the synthetic ratio to 50\% yields a further improvement to 79.8 mAP, the highest multi-spectral result across all configurations, though at the cost of overall camouflaged performance dropping to 64.9 mAP, reflecting a trade-off between multi-spectral specialization and general robustness.

\subsubsection{Non-generative baseline: black-bar augmentation}
The black-bar augmentation baseline improves overall camouflaged detection performance from 56.8 to 62.6 mAP. However, it remains well below Qwen zero-shot (69.9 mAP), suggesting that realistic camouflage appearance contributes additional benefit beyond partial occlusion alone. This difference is particularly evident for foliage and netting camouflage, where Qwen zero-shot exceeds the black-bar baseline by 15.6 mAP and 6.3 mAP, respectively. In contrast, performance on multi-spectral camouflage is comparable between black-bar augmentation and Qwen zero-shot (71.2 versus 70.9 mAP).

\section{Discussion}
\label{sec:discussion}
The results demonstrate that synthetic camouflage augmentation can meaningfully reduce a domain gap that would otherwise require real camouflaged training data. Real uncamouflaged images anchor the detector to accurate vehicle appearance, while synthetic camouflage exposes it to target-domain variation it would never encounter during standard training, so the two sources provide complementary rather than competing information. This improvement comes at no cost to in-domain performance: under the best-performing 25\% synthetic data configuration, detection performance on uncamouflaged imagery remains stable across augmentation approaches. Instead, differences between approaches are primarily determined by the quality of the synthetic camouflage. 

The magnitude of the benefit depends on the specific camouflage type. Relative to the uncamouflaged baseline, the largest performance drops are observed for foliage and netting camouflage, as both introduce physical occlusions that break up the vehicle silhouette and disrupt the edge and shape cues that detectors rely on for feature extraction. Multi-spectral camouflage results in a comparatively smaller decrease of around 11 mAP, as it alters surface texture without significantly modifying overall shape, leaving structural cues such as edges, contours, and distinctive geometries largely intact.

The black-bar augmentation shows that part of the performance gain does not depend on realistic camouflage synthesis. Despite producing visually implausible occlusions, the black-bar augmentation improves overall camouflaged detection performance from 56.8 to 62.6 mAP. This suggests that a substantial portion of the gain arises simply from exposing the detector to partial occlusion. However, the effect of occlusion appears to be largely captured by the generative augmentation itself. Combining Qwen-generated camouflage with black-bar augmentation does not improve overall performance beyond Qwen alone, with both configurations achieving 69.9 mAP. 

Furthermore, realistic synthetic camouflage provides additional benefit when it accurately reflects the target domain. Qwen exceeds the black-bar augmentation by 6.3 mAP on netting and 15.6 mAP on foliage. In contrast, no such advantage is observed for multi-spectral camouflage, where zero-shot generation fails to accurately reproduce the target appearance. These results suggest that realism is not required to obtain a benefit from augmentation, but it is critical for reducing the remaining domain gap beyond what can be achieved through simple occlusion alone.

LoRA fine-tuning improves detection performance for the camouflage type that zero-shot editing cannot reliably represent. Visual inspection indicates that both zero-shot Qwen and Flux reproduce foliage and netting camouflage plausibly, but neither reliably captures the distinctive appearance of multi-spectral camouflage from text prompts alone. This is likely because foliage and netting are well represented in the models' pretraining data, whereas multi-spectral camouflage is a specialized military concept that is largely absent from general-domain image collections. Fine-tuning on a small set of synthetic before (uncamouflaged) and real after (multi-spectral camouflaged) image pairs substantially improves multi-spectral detection performance, indicating that the zero-shot performance gap was driven primarily by insufficient target-domain representation rather than a fundamental limitation of the image-editing approach itself. In contrast, fine-tuning provides no consistent benefit for foliage and netting, where zero-shot editing already produces sufficiently representative camouflage. The practical implication is that zero-shot editing appears sufficient when the target appearance is well represented in the base model, whereas fine-tuning becomes valuable when domain-specific appearances fall outside the model's existing visual knowledge.

The performance difference between Qwen and Flux highlights the importance of synthetic data quality. Across nearly all camouflage conditions, Qwen produces larger detector performance gains and lower variance than Flux. Visual inspection suggested that Qwen better preserves vehicle geometry while generating camouflage that more closely matches the target-domain appearance. In contrast, Flux frequently exhibits weaker prompt adherence, producing camouflage that is less consistent with the intended target appearance. This may explain both the lower mean performance and the higher run-to-run variability observed when training on Flux-generated data.

\subsection{Limitations}
The prompt engineering process relies on qualitative visual assessment and iterative experimentation rather than a systematic optimization procedure. The final prompts reflect the best configurations identified during development, but alternative formulations may produce more realistic or diverse outputs. This subjectivity extends to the LoRA fine-tuning stage, where prompts required manual adjustment to account for differences in model behaviour.

A second limitation concerns the use of non-camouflaged images for checkpoint selection of the detection models. While this makes sense for a baseline model, it is suboptimal to select the model that works best on camouflaged images. An alternative would have been to use the same number of iterations for each experiment, without early stopping.

The camouflaged test set is limited in size and diversity. The number of images per vehicle class and camouflage type varies, and certain combinations are underrepresented due to the scarcity of publicly available imagery. Although this limitation primarily affects the reliability of the evaluation, rather than the proposed augmentation approach, a larger and more balanced test set would enable a more robust and fine-grained evaluation.

Finally, all detection results are obtained with a single detector, MM GroundingDINO with a Swin-T backbone. The improvements from synthetic camouflage augmentation are therefore established for this architecture and may not transfer directly to detectors with different inductive biases. 

\subsection{Conclusion}
This work demonstrates that generative image editing is a practical and scalable approach to reducing domain gaps in low-data regimes. Zero-shot editing suffices when the generative model has adequate prior knowledge of the target appearance, while fine-tuning on a small set of domain-specific examples closes remaining gaps when it does not. Together, these results show that the effectiveness of synthetic augmentation depends strongly on its fidelity to the target appearance. When synthetic camouflage adequately represents the target domain, generative augmentation consistently outperforms non-generative augmentation and provides an effective mechanism for improving detector robustness in low-data settings.

\bibliography{report} 
\bibliographystyle{spiebib} 
\section{Supplementary - Extended results}
\label{supplementary:performance}

\subsection{Effect of synthetic data ratio}
Across all pipelines, increasing the proportion of synthetic data beyond 25–50\% generally degrades performance, with the effect most severe for Flux.2 Dev. At 100\% synthetic data, Flux.2 Dev drops to 30.0 mAP on the camouflaged test set, and to 28.4 mAP on uncamouflaged images. This occurs because training exclusively on synthetic images causes the detector to overfit to generation artifacts that are absent from real images, so the learned features fail to transfer at test time. For Qwen, the degradation is far more gradual: at 100\% synthetic it still achieves 54.9 mAP, suggesting its generated images are substantially more consistent with the real data distribution. This pattern is illustrated in Table~\ref{fig:results_camo_uncamo} and Table~\ref{fig:results_extended_camos}. One exception to this degradation pattern is black-bar augmentation on multi-spectral camouflage, which peaks at 50\% synthetic data (77.3 mAP), slightly exceeding the uncamouflaged baseline of 75.4 mAP.

\begin{table}[b] \centering \caption{Performance across Camouflage and Uncamouflage Test Sets. Mean Average Precision (mAP@50--95) for models trained with varying mixtures of real (uncamouflaged) and synthetic (camouflaged) data (0\%, 25\%, 50\%, 75\%, and 100\% synthetic). The combined experiment with black-bar augmentation and Qwen was evaluated only up to the 50\% ratio; dashes (\textemdash) indicate ratios not run.} \label{tab:comparison_overall} 
\tiny
\begin{subtable}{\textwidth} \centering \caption{Evaluation on the \textbf{Camouflage} Test Set} \resizebox{\textwidth}{!}{ \begin{tabular}{l | ccccc} \toprule Method & $0\uparrow$ & $25\uparrow$ & $50\uparrow$ & $75\uparrow$ & $100\uparrow$ \\ \midrule Flux.2 Dev & 56.8$\pm$1.8 & 61.4$\pm$5.2 & 55.5$\pm$3.1 & 48.6$\pm$2.6 & 30.0$\pm$2.9 \\ Flux.2 Dev Blurred & 56.8$\pm$1.8 & 59.5$\pm$3.1 & 54.9$\pm$2.6 & 49.3$\pm$2.4 & 31.4$\pm$1.4 \\ Flux.2 Dev second pass & 56.8$\pm$1.8 & 54.9$\pm$4.2 & 43.1$\pm$0.5 & 52.2$\pm$1.5 & 34.0$\pm$2.5 \\ Flux.2 Dev Filtered & 56.8$\pm$1.8 & 58.4$\pm$3.2 & 58.3$\pm$4.3 & 51.6$\pm$3.1 & 27.0$\pm$2.0 \\ \midrule Qwen Image Edit 2509 & 56.8$\pm$1.8 & \textbf{69.9}$\pm$\textbf{2.1} & 61.5$\pm$3.2 & 63.8$\pm$1.4 & 54.9$\pm$1.4 \\ Qwen Image Edit 2509 Filtered & 56.8$\pm$1.8 & 67.1$\pm$2.9 & 65.4$\pm$5.0 & 60.5$\pm$1.4 & 60.0$\pm$3.8 \\ Qwen Image Edit 2509 fine-tuned & 56.8$\pm$1.8 & 67.5$\pm$2.6 & 64.9$\pm$2.6 & 60.3$\pm$3.1 & 60.5$\pm$1.9 \\ Qwen Image Edit 2509 fine-tuned filtered & 56.8$\pm$1.8 & 63.8$\pm$2.6 & 65.2$\pm$3.8 & 65.5$\pm$2.6 & 60.2$\pm$2.5 \\ \midrule Black-bar augmentation & 56.8$\pm$1.8 & 62.6$\pm$1.8 & 62.8$\pm$2.7 & 57.8$\pm$2.4 & 54.7$\pm$3.3 \\ \midrule Black-bar augmentation \& Qwen & 56.8$\pm$1.8 & 69.9$\pm$2.4 & 66.3$\pm$1.5 & \textemdash & \textemdash \\ \bottomrule \end{tabular} } \end{subtable} \vspace{1em} 
\begin{subtable}{\textwidth} \centering \caption{Evaluation on the \textbf{Uncamouflaged} Test Set} \resizebox{\textwidth}{!}{ \begin{tabular}{l | ccccc} \toprule Method & $0\uparrow$ & $25\uparrow$ & $50\uparrow$ & $75\uparrow$ & $100\uparrow$ \\ \midrule Flux.2 Dev & 86.6$\pm$2.5 & 87.3$\pm$1.6 & 85.4$\pm$3.6 & 75.9$\pm$8.8 & 28.4$\pm$8.2 \\ Flux.2 Dev Blurred & 86.6$\pm$2.5 & 86.1$\pm$2.5 & 85.9$\pm$2.1 & 79.3$\pm$6.8 & 32.8$\pm$8.2 \\ Flux.2 Dev second pass & 86.6$\pm$2.5 & 86.5$\pm$3.1 & 87.0$\pm$2.2 & 81.6$\pm$2.8 & 31.1$\pm$9.2 \\ Flux.2 Dev Filtered & 86.6$\pm$2.5 & 86.4$\pm$3.0 & 84.0$\pm$4.0 & 80.7$\pm$4.1 & 25.9$\pm$7.4 \\ \midrule Qwen Image Edit 2509 & 86.6$\pm$2.5 & 87.7$\pm$2.3 & 85.2$\pm$3.5 & 81.9$\pm$4.0 & 72.4$\pm$6.2 \\ Qwen Image Edit 2509 Filtered & 86.6$\pm$2.5 & 82.1$\pm$5.3 & 84.3$\pm$5.3 & 80.9$\pm$3.7 & 76.6$\pm$4.6 \\ Qwen Image Edit 2509 fine-tuned & 86.6$\pm$2.5 & 86.1$\pm$3.0 & 85.6$\pm$3.2 & 80.9$\pm$3.9 & 76.1$\pm$3.2 \\ Qwen Image Edit 2509 fine-tuned filtered & 86.6$\pm$2.5 & 83.1$\pm$3.2 & 85.1$\pm$3.0 & 85.0$\pm$2.9 & 76.5$\pm$4.7 \\ \midrule Black-bar augmentation & 86.6$\pm$2.5 & 87.5$\pm$1.8 & 86.2$\pm$4.4 & 86.4$\pm$3.5 & 77.0$\pm$6.5 \\ \midrule Black-bar augmentation \& Qwen & 86.6$\pm$2.5 & 85.6$\pm$2.9 & 83.8$\pm$0.9 & \textemdash & \textemdash \\ \bottomrule \end{tabular} } \end{subtable} \label{fig:results_camo_uncamo} \end{table}

\subsubsection{Effect of quality filtering and post-processing}
A closer inspection of the Flux-generated data shows that the added camouflage tends to be rendered sharper than the original image it is applied to, introducing a statistical discrepancy that could lead the detector to key on editing artifacts rather than camouflage-relevant features. Two post-processing strategies were evaluated to suppress these discrepancies. Gaussian blurring corrects the sharpness mismatch, while a second diffusion pass harmonizes lower-level noise, texture, and color inconsistencies (Section~\ref{methods:postprocessing}). Neither yields a consistent improvement at 25\% synthetic data, the best-performing ratio (Table~\ref{fig:results_camo_uncamo}). Gaussian blurring produces no meaningful change relative to the unprocessed Flux baseline (59.5 vs.\ 61.4 mAP), and the second diffusion pass degrades performance to 54.9 mAP, below the uncamouflaged baseline of 56.8 mAP, with no visible difference between processed and unprocessed images on inspection. These results indicate that the discrepancies these strategies target are not the primary source of the Flux and Qwen performance gap. The effect is similarly inconsistent across the remaining synthetic ratios: the only marginal gains appear at high synthetic ratios where overall performance is already degraded. Quality filtering, with approximately 10\% of generated images flagged as faulty in each case, had no meaningful effect on detection across any generative pipeline.

\begin{table}[b] \centering \caption{Performance across different Camouflage Test Sets. Mean Average Precision (mAP@50--95) for models trained with varying mixtures of real (uncamouflaged) and synthetic (camouflaged) data (0\%, 25\%, 50\%, 75\%, and 100\% synthetic).} \label{tab:comparison_overall_specific} \begin{subtable}{\textwidth} \centering
\tiny
\caption{Evaluation on the \textbf{foliage} camouflage test set} \label{tab:comparison_recallmAP_foliage} \resizebox{\textwidth}{!}{ \begin{tabular}{l | ccccc} \toprule Method & $0\uparrow$ & $25\uparrow$ & $50\uparrow$ & $75\uparrow$ & $100\uparrow$ \\ \midrule Flux.2 Dev & 54.1$\pm$1.0 & 59.1$\pm$3.3 & 51.9$\pm$3.1 & 45.2$\pm$5.0 & 32.9$\pm$4.8 \\ Flux.2 Dev second pass & 54.1$\pm$1.0 & 47.9$\pm$3.0 & 40.6$\pm$1.1 & 34.2$\pm$2.9 & 30.0$\pm$1.7 \\ Flux.2 Dev Filtered & 54.1$\pm$1.0 & 60.5$\pm$4.7 & 55.1$\pm$1.6 & 48.1$\pm$2.1 & 50.0$\pm$4.3 \\ \midrule Qwen Image Edit 2509 & 54.1$\pm$1.0 & 74.2$\pm$1.6 & 66.5$\pm$3.6 & 65.5$\pm$2.9 & 57.7$\pm$3.0 \\ Qwen Image Edit 2509 Filtered & 54.1$\pm$1.0 & 67.9$\pm$5.4 & 66.7$\pm$6.8 & 61.5$\pm$1.6 & 59.5$\pm$8.1 \\ \midrule Qwen Image Edit 2509 fine-tuned & 54.1$\pm$1.0 & 68.9$\pm$1.4 & 62.9$\pm$3.8 & 58.0$\pm$4.0 & 60.0$\pm$0.9 \\ Qwen Image Edit 2509 fine-tuned filtered & 54.1$\pm$1.0 & 63.7$\pm$3.6 & 66.9$\pm$1.0 & 64.0$\pm$1.9 & 63.0$\pm$2.8 \\ \midrule Black-bar augmentation & 54.1$\pm$1.0 & 58.6$\pm$2.5 & 61.5$\pm$0.9 & 53.6$\pm$2.0 & 47.8$\pm$6.8 \\ \bottomrule \end{tabular} } \end{subtable} \vspace{1em} \begin{subtable}{\textwidth}
\tiny
\centering \caption{Evaluation on the \textbf{netting} camouflage test set} \label{tab:comparison_recallmAP_net} \resizebox{\textwidth}{!}{ \begin{tabular}{l | ccccc} \toprule Method & $0\uparrow$ & $25\uparrow$ & $50\uparrow$ & $75\uparrow$ & $100\uparrow$ \\ \midrule Flux.2 Dev & 60.4$\pm$3.3 & 70.4$\pm$4.7 & 65.6$\pm$4.5 & 57.2$\pm$1.9 & 32.2$\pm$4.6 \\ Flux.2 Dev second pass & 60.4$\pm$3.3 & 62.2$\pm$3.1 & 51.4$\pm$2.2 & 46.0$\pm$2.9 & 41.8$\pm$1.5 \\ Flux.2 Dev Filtered & 60.4$\pm$3.3 & 65.0$\pm$2.8 & 68.4$\pm$4.6 & 60.2$\pm$5.9 & 59.7$\pm$2.2 \\ \midrule Qwen Image Edit 2509 & 60.4$\pm$3.3 & 74.8$\pm$0.8 & 67.9$\pm$2.1 & 70.0$\pm$4.3 & 60.9$\pm$1.9 \\ Qwen Image Edit 2509 Filtered & 60.4$\pm$3.3 & 73.5$\pm$3.6 & 70.5$\pm$7.1 & 66.4$\pm$2.7 & 67.8$\pm$3.0 \\ \midrule Qwen Image Edit 2509 fine-tuned & 60.4$\pm$3.3 & 74.3$\pm$3.2 & 72.3$\pm$2.6 & 70.4$\pm$2.9 & 67.7$\pm$3.1 \\ Qwen Image Edit 2509 fine-tuned filtered & 60.4$\pm$3.3 & 70.5$\pm$2.1 & 71.4$\pm$5.4 & 72.1$\pm$1.6 & 66.6$\pm$3.5 \\ \midrule Black-bar augmentation & 60.4$\pm$3.3 & 68.5$\pm$4.2 & 66.2$\pm$3.1 & 63.3$\pm$2.2 & 55.5$\pm$6.0 \\ \bottomrule \end{tabular} } \end{subtable} \vspace{1em} \begin{subtable}{\textwidth}
\tiny
\centering \caption{Evaluation on the \textbf{multi-spectral} camouflage test set} \label{tab:comparison_recallmAP_multispectral} \resizebox{\textwidth}{!}{ \begin{tabular}{l | ccccc} \toprule Method & $0\uparrow$ & $25\uparrow$ & $50\uparrow$ & $75\uparrow$ & $100\uparrow$ \\ \midrule Flux.2 Dev & 75.4$\pm$2.6 & 66.5$\pm$5.7 & 65.6$\pm$0.8 & 62.0$\pm$2.0 & 57.3$\pm$1.6 \\ Flux.2 Dev second pass & 75.4$\pm$2.6 & 65.6$\pm$0.8 & 53.0$\pm$5.2 & 51.8$\pm$2.1 & 49.4$\pm$3.2 \\ Flux.2 Dev Filtered & 75.4$\pm$2.6 & 68.7$\pm$3.7 & 69.2$\pm$8.0 & 64.6$\pm$4.0 & 65.4$\pm$3.4 \\ \midrule Qwen Image Edit 2509 & 75.4$\pm$2.6 & 70.9$\pm$4.6 & 67.1$\pm$1.4 & 68.2$\pm$1.9 & 65.1$\pm$1.9 \\ Qwen Image Edit 2509 Filtered & 75.4$\pm$2.6 & 70.5$\pm$2.0 & 73.4$\pm$2.9 & 66.8$\pm$1.6 & 64.7$\pm$2.6 \\ \midrule Qwen Image Edit 2509 fine-tuned & 75.4$\pm$2.6 & 76.5$\pm$6.4 & 79.8$\pm$1.4 & 70.2$\pm$3.2 & 73.3$\pm$3.5 \\ Qwen Image Edit 2509 fine-tuned filtered & 75.4$\pm$2.6 & 76.4$\pm$4.8 & 76.6$\pm$1.9 & 76.4$\pm$3.1 & 71.7$\pm$0.8 \\ \midrule Black-bar augmentation & 75.4$\pm$2.6 & 71.2$\pm$0.7 & 77.3$\pm$5.5 & 70.5$\pm$7.8 & 69.7$\pm$2.9 \\ \bottomrule \end{tabular} } \end{subtable} \label{fig:results_extended_camos} \end{table}

\end{document}